# Evaluation of Semantic Web Technologies for Storing Computable Definitions of Electronic Health Records Phenotyping Algorithms

Václav Papež[1,2,*], MSc, Spiros Denaxas[1,2,*], PhD, Harry Hemingway[1,2], FRCP
[1] Institute of Health Informatics, University College London, London, UK
[2] Farr Institute of Health Informatics Research, University College London, London, UK

**Abstract**

*Electronic Health Records are electronic data generated during or as a byproduct of routine patient care. Structured, semi-structured and unstructured EHR offer researchers unprecedented phenotypic breadth and depth and have the potential to accelerate the development of precision medicine approaches at scale. A main EHR use-case is defining phenotyping algorithms that identify disease status, onset and severity. Phenotyping algorithms utilize diagnoses, prescriptions, laboratory tests, symptoms and other elements in order to identify patients with or without a specific trait. No common standardized, structured, computable format exists for storing phenotyping algorithms. The majority of algorithms are stored as human-readable descriptive text documents making their translation to code challenging due to their inherent complexity and hinders their sharing and re-use across the community. In this paper, we evaluate the two key Semantic Web Technologies, the Web Ontology Language and the Resource Description Framework, for enabling computable representations of EHR-driven phenotyping algorithms.*

## Introduction

Electronic Health Records (EHR) are structured, semi-structured and unstructured data that are generated during routine interactions of patients with primary care, hospital care and tertiary healthcare or as a byproduct of those interactions for billing or administrative purposes[1]. Structured EHR are recorded using controlled clinical terminologies while unstructured data include clinical text and narrative. Semi-structured EHR data often loosely follow a data specification (e.g. prescription events, medical imaging reports) but this varies greatly across information systems, clinical specialties and healthcare providers. High-throughput genotyping and increased availability of EHR data are giving scientists the unprecedented opportunity to exploit routinely generated clinical data to advance precision medicine at scale. EHR data can fundamentally alter the manner in which genetic association studies are performed and enable scientists to examine the association of genetic variants and traits in larger sample sizes and phenotypic breadth[2].

A primary use-case of EHR data is the creation of phenotyping (or "case finding") algorithms[3], computational algorithms that identify patients that have (or have not) been diagnosed with a particular condition[4] (e.g. acute myocardial infarction, prostate cancer, or anxiety etc.) and where applicable the disease onset and severity. Phenotyping algorithms tend to use clinical information such as diagnoses, laboratory tests, symptoms, clinical examination findings, prescriptions, referrals and other EHR data elements. While the term *phenotype* is traditionally defined as the physical manifestation of a particular trait, in the context of EHR research, phenotypes are broadly (but not exclusively) as the presence or absence of a particular clinical condition. In EHR resources linked with genetic data, such as the Electronic Medical Records and Genomics (eMERGE) consortium[5], these phenotypes can enable large-scale genomic association studies which have been traditionally limited to a small set of traits. Phenotyping however is a challenging and time-consuming process since often data been collected for care, auditing or administrative purposes and not for research. The contents of EHR data sources are an indirect representation of the true patient state as skewed by the underlying healthcare process e.g. clinical guidelines, information systems, data standards[6].

Defining and validating EHR phenotyping algorithms is challenging and time-consuming. Challenges are amplified by the lack of a common definition standard for algorithms, making their sharing across the scientific community problematic. Despite the fact that phenotype components are structured and often annotated by controlled clinical terminology terms, phenotype definitions, and their underlying logic are usually expressed as free-text which is not readily machine-readable. The translation from this narrative to programming code used to identify and extract patients (e.g. implementing a phenotyping algorithm using Structured Query Language for use in a relational database management system) can be problematic due to potential ambiguities in the manner in which the algorithm was expressed or potential ways of implementing it using local data. There is a clear and urgent need to develop and evaluate a computable, standards-driven format to facilitate the systematic creation, sharing and re-use of EHR-derived phenotyping algorithms.

*Semantic Web Technologies (SWT)*

Semantic Web Technologies are a collection of World Wide Web Consortium (W3C) standards (https://www.w3.org) for annotating and sharing data using Web protocols and can potentially address some of these challenges and be utilized to define computable EHR-derived phenotypes. A key advantage of SWT is that they were specifically built to facilitate the automated integration and reuse of data in a machine-readable manner while in parallel enabling standard Web-driven user interaction in human-friendly formats. The Ontology Web Language (OWL) (https://www.w3.org/OWL/) is a specification of an ontology language based on a description logic. An ontology in OWL is mostly modeled as an RDF (Resource Description Framework) graph (https://www.w3.org/RDF/), a specification to describe data/resources in form of *triples* <subject-predicate-object>, where subject and predicate are resources and object is also a resource or a literal (value). Resources, uniquely identified by Internationalized Resource Identifiers (IRIs) which in turn are a generalization of Uniform Resource Identifiers (URIs), represent nodes (subjects and objects) and edges (relationships) of the graph structure and leaf nodes are created by literals (values) as for example in Figure 3.

The RDF Schema (RDFS) provides a set of basic predefined semantics (i.e. resources) for RDF graphs including data types and elements for the construction of classes and subclasses. OWL provides extended semantic expressions and constructs such as set operations, class disjoints, class equivalencies, and cardinality constraints. Semantic expressiveness (and thus a set of OWL constructs) is supplied by the OWL dialect and profile and influence the behavior of the *semantic reasoners*, engines able to automatically infer consequences and associations within the ontology which are not explicitly specified. The querying of RDF is enabled by the SPARQL Protocol and RDF Query Language (SPARQL) semantic language. SPARQL (https://www.w3.org/TR/rdf-sparql-query/) has a similar structure to Structured Query Language (SQL) which is widely used in relational database systems and enables queries for data stored as RDF through query processing engines called SPARQL end-points.

The use of SWT has been previously evaluated in the field of EHR phenotyping[7]. The work by Pathak and colleagues was focused on the transformation of EHR data from multiple heterogeneous sources into RDF graphs. The created RDF graphs which were then queried via a single SPARQL end-point in order to assess research cohort. The research outlined in this paper is complementary to their work in that it is focused on the explicit description of the phenotype algorithm itself. EHRs are included into the process in the second step and outcome cohorts (or their parts) are inferred automatically by semantic reasoners. Compared to previous work, our approach describes how a phenotype algorithm and its logic can be defined by OWL constructs and queries can be implemented in SPARQL.

In this paper, we implement an OWL/RDF ontology-based approach for storing a deterministic EHR-derived diabetes phenotyping algorithm. We validate our approach against a pre-defined list of desiderata developed previously. We used diabetes as a case study since it exemplifies many of the associated challenges but our findings are generalizable to other diseases and syndromes.

**Methods**

*CALIBER*

We used a deterministic EHR phenotyping algorithm for diabetes developed and validated in the CALIBER resource[8]. CALIBER is a translational research platform which links national, structured primary care, hospital care, disease registry, mortality data and socioeconomic information in the UK for ~10m patients. Primary care data are provided by the Clinical Practice Research Datalink, an anonymized national cohort of longitudinal data for all individuals registered with a general practitioner. Secondary care data are recorded in Hospital Episode Statistics (HES), a national database of administrative data used for hospital reimbursement. Finally, mortality and socioeconomic data are collected and curated by the Office of National Statistics (ONS).

Primary care data include diagnoses, referrals, symptoms, laboratory tests and clinical examination findings recorded using the Read controlled clinical terminology (a subset of SNOMED-CT, http://www.snomed.org/snomed-ct). Medication prescriptions are organized using the British National Formulary (https://www.bnf.org/), a structured resource for classifying all therapeutic agents prescribed in UK healthcare. Hospital care data include ranked diagnoses recorded using ICD-10 terms and interventional procedures recorded using the OPCS Classification of Interventions and Procedures version 4 (OPCS4) terms which are essentially semantically equivalent to the Current Procedural

Terminology (CPS) terms used in the United States. Mortality data are recorded using ICD-9 and ICD-10 and include the underlying cause of death and up to 15 contributory causes of mortality.

*Diabetes phenotyping algorithm*

Diabetes affects approximately 25 million people in the US and is one of the most common non-communicable diseases globally associated with a significant burden to patients and healthcare systems. Diabetes has a number of environmental (e.g. body mass index) and genetic risk factors and is one of the leading causes of heart disease and stroke. In addition, diabetes complications such as neuropath, nephropathy, retinopathy and amputation have a substantial impact of the quality of life of patients. The ability of researchers to undertake studies in diabetes using electronic health records potentially linked to genetic data is of high value.

We used a previously validated[9] EHR-derived phenotyping algorithm for identifying and classifying patients with diabetes into four distinct, non-overlapping groups: 1) patients with type 1 diabetes, 2) patients with type 2 diabetes, 3) patients with unspecified diabetes and 4) patients that are not diabetic. The algorithm (Figure 1) combines clinical information from specific diagnostic codes for type 1 and type 2 diabetes with less specific codes for 'insulin dependent diabetes' (IDDM) and 'non-insulin dependent diabetes' (NIDDM) recorded across both primary and secondary care. Patients with diagnoses of both type 1 and type 2 diabetes are classified as patients with diabetes of unspecified type. The algorithm was designed to primarily identify patients with type 2 diabetes that can be on a variety of medications so it does not make use of medication data explicitly. Further refinements would be required to identify patients with type 1 diabetes with greater recall.

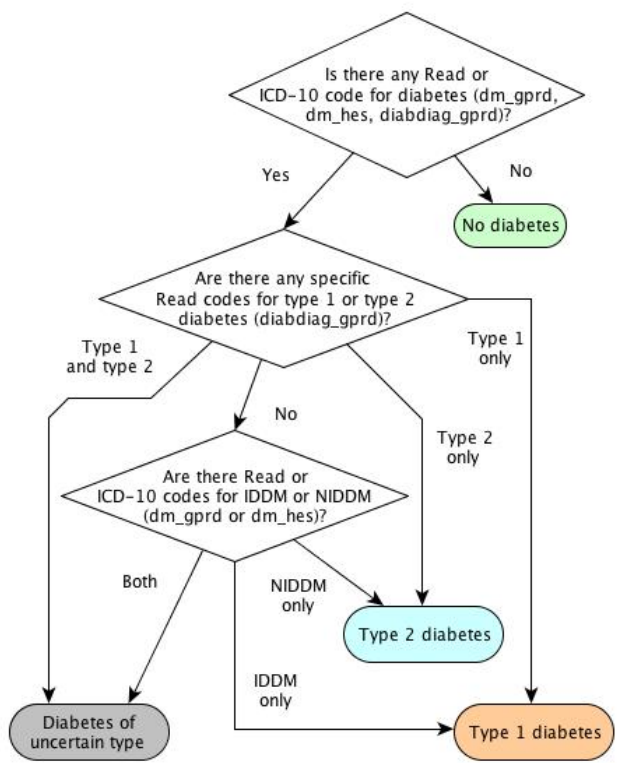

**Figure 1.** CALIBER diabetes phenotyping algorithm: patients are classified in four distinct categories: type 1 diabetes, type 2 diabetes, diabetes of uncertain type and non-diabetic on the basis of diagnostic codes found in primary care EHR or hospital administrative data.

Individual phenotype components are stratified by data source - e.g. *dm_gprd* represents a diagnosis of diabetes from primary care, *dm_hes* a diagnosis from secondary care. Within each component, diagnostic terms from controlled clinical terminologies have been grouped by clinicians in terms of certainty e.g. *historical diagnoses, possible* *diagnosis, confirmed diagnosis*. All individual components and definitions are made available through an open-access Data Portal (https://www.caliberresearch.org/portal/show/phenotype_diabetes).

*Desiderata*

We used the desiderata defined by Mo and colleagues[10] in order to evaluate RDF/OWL for storing computable representations of phenotyping algorithms. In their work, the authors reviewed a series of EHR phenotyping algorithms (e.g. dementia, Crohn's disease, cataract, HDL) developed as part of eMERGE and a series of authoring tools such as the Measure Authoring Tool (MAT) (https://www.emeasuretool.cms.gov/), i2b2 (https://www.i2b2.org/), and the SHARPn PhenotypePortal (http://phenotypeportal.org/). The authors propose a list of recommendations for desired features (*desiderata*) for computable phenotype representations models:

i. Support both human-readable and computable representations

ii. Implement set operations and relational algebra

iii. Represent phenotype criteria using structured rules (e.g. nested logical structure, Boolean logic, comparative operations, aggregative operations, and negation)

iv. Support defining temporal relations between clinical events

v. Utilize standardized controlled clinical terminologies and facilitate reuse of value sets

vi. Provide interfaces for external software algorithms or data components such as support for Natural Language Processing (NLP) operations for extracting clinically significant markers from unstructured EHR

vii. Maintain backwards compatibility to accommodate for temporal changes in the underlying healthcare process model and EHR data specifications.

**Results**

*Phenotype ontology*

As part of our work, a proof of concept application (Figure 2) for generating OWL phenotype ontologies was developed in Java using the OWL API, the Hermit reasoner (http://www.hermit-reasoner.com/) and the Protégé, software (http://protege.stanford.edu/). Phenotype ontologies were serialized into OWL/XML format by the semantic reasoners. We utilized SPARQL for defining and performing queries and additional operations, but the main algorithmic logic is defined by OWL constructs.

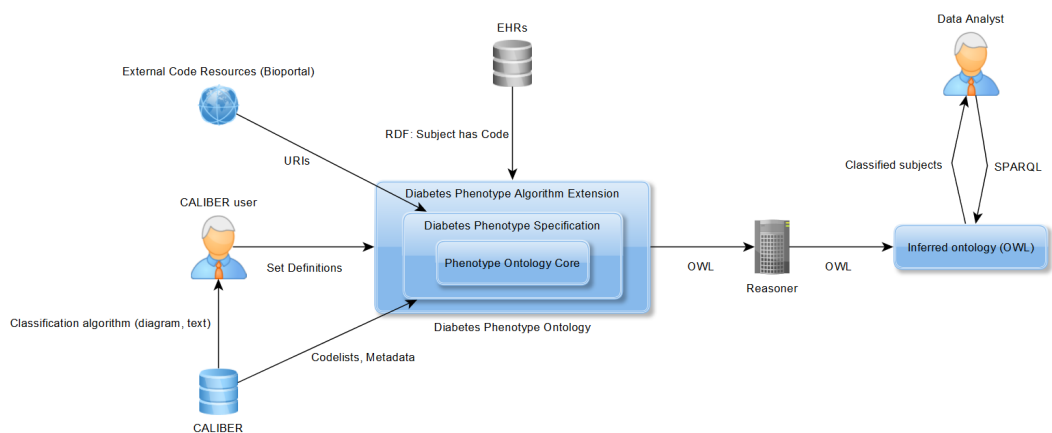

**Figure 2.** Diagram of the ontology-based phenotyping solution

We developed the following process to incrementally generate and extend the phenotype ontology using OWL DL:

1) **Generic structural core**: A generic ontology structure is predefined which essentially defines the generic structure for phenotyping algorithms.

2) **Phenotype components**: Individual components of phenotypes, such as lists of diagnostic terms associated with a diagnosis of diabetes in primary care or in administrative data in hospitals is processed. This results in the creation of phenotype element classes (i.e. component, category, diagnostic term) that will hold this information. Diagnostic term lists (*code lists*) associated with phenotype components (e.g. diagnoses of type 2 diabetes in primary care) are currently stored using a very simple bespoke format in CALIBER which enables their automatic processing and addition to the ontology as new classes and individuals. Terms from controlled clinical terminologies used in the algorithm specified as the literals can be defined using IRIs from external ontology/terminology repositories such as the Bioportal, the repository of biomedical ontologies (http://bioportal.bioontology.org/), which provides ontologies derived from e.g. SNOMED-CT, Read or ICD-10/9 vocabularies.

3) **Phenotype algorithm logic**: In this stage, the logic associated with a phenotyping algorithm is manually specified enabling the algorithm to be decomposed into sets and set operations. The phenotype algorithm is based on narrowing the initial set of patients to a final subset representing patients with type 1 diabetes, type 2 diabetes, and unknown type diabetes. Therefore, the algorithm can be described as a graph where nodes represent sets of patients (classes from the ontology perspective) and edges represent set operations and constraints. For example, a class defined as *subjects which have asserted at least 1 Read code of type 1 diabetes intersected with subjects which have asserted at least 1 Read code of type 2 diabetes* is a subclass of class of *subjects with unknown type diabetes*. Union of all subclasses of *unknown type diabetes* is equivalent to *subject with unknown type diabetes* class. With this step, the OWL phenotype description can be serialized. Class names are generated automatically as a composition of the original element names as they are presented in CALIBER and suffixes (and/or prefixes) of general class names.

4) **Subject classification**: In this final stage, structured EHRs are automatically transformed into RDF triples <patient-obtained-code> and added to the graph (Figure 3). Individual patients are asserted to the class *Subject* and individual diagnostic terms are asserted to the class *Code*. As the code and patient id are unique, Internationalized Resource Identifiers (IRIs) of those individuals are concatenated from ontology IRI and patient's id/code. This step does not describe the phenotype itself, but it uses the phenotype for subject classification.

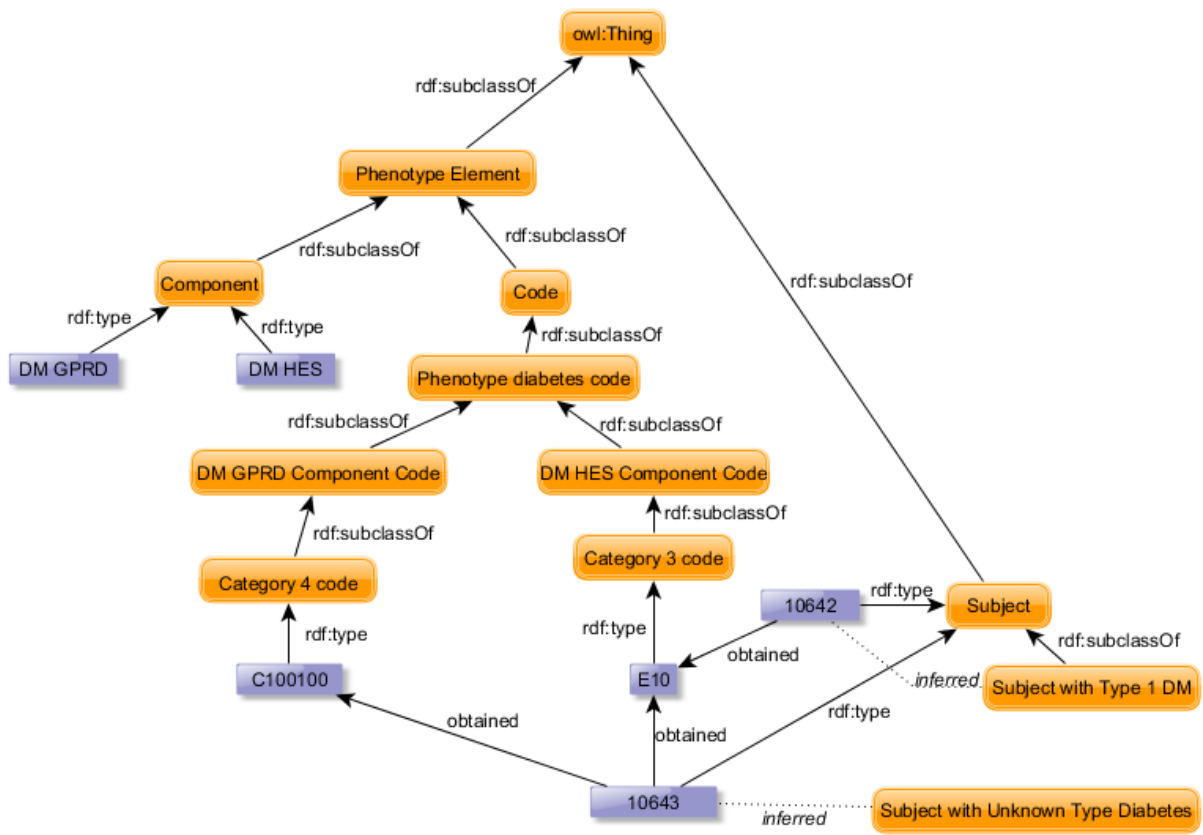

**Figure 3.** Part of the phenotype ontology including two imported subjects (patient pseudo-identifiers 10642 and 10643) and consequence inferences as generated from the reasoner, generated using Protégé 5. C100100 is the Read identifier for the term "Diabetes mellitus, adult onset, no mention of complication" and E10 is the ICD-10 term for "Type 1 diabetes mellitus". DM GPRD is diagnosis of diabetes in primary care, DM HES is a diagnosis of diabetes in hospital care. Category 3 is "Insulin dependent diabetes" and Category 4 is "Non-insulin dependent diabetes". Algorithm implementation details available on CALIBER Data Portal: https://www.caliberresearch.org/portal/show/dm_hes.

Once these steps are complete, the automatic semantic reasoner (e.g. Hermit, Fact++) can be applied over the ontology and new individual assertion axioms are inferred and stored in separate RDF graphs. From these inferences (Figure 4 shows inferred assertions of patients contained within a yellow rectangle), individual groups of classified patients (*cohorts*) can be extracted using SPARQL queries in which the additional constraints can be specified if required. OWL supports constructs specifying constraints as disjoint classes, complement classes, class closures or zero individual occurrence, but semantic reasoners operate under *an open world assumptions* e.g. they assume the truth value may be true irrespective of whether or not it is *known* to be true. As a result, to reach the exact patients subset which is constrained by non-existence of some elements, e.g. patients obtained no diagnostic term from secondary health care, additional restrictions into SPARQL queries has to be added. As an example, the semantic reasoner can infer all patients that have a diabetes type 1 diagnostic term and all patients that have both diabetes type 1 and type 2 diagnostic terms. Patients with only diabetes type 1 diagnostic terms can be query as an intersection of the previously mentioned classes and this intersection can be easily specified within the WHERE clause of SPARQL (Figure 6).

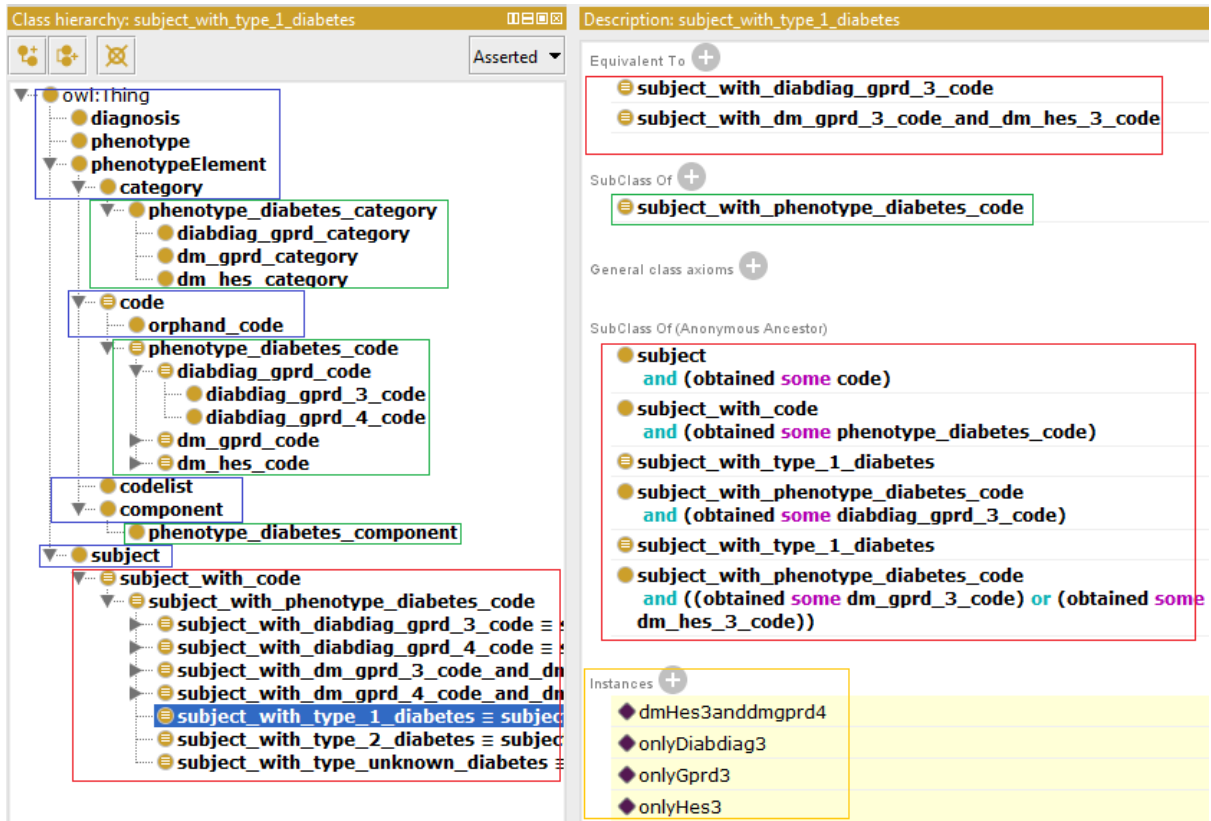

**Figure 4**. Part of the diabetes phenotype ontology. Blue boxes represent the core; green boxes represent the automatically imported elements; red boxes represent manually added elements; yellow box shows the inferred instances, which are otherwise asserted to the class subject

*(i) Support both human-readable and computable representations*

OWL/RDF directly supports various machine-readable serialization formats with XML-based serialization into OWL/XML (RDF/XML respectively) being the most common approach. A major advantage is the implicit compatibility with common XML parsers for integrating into external applications. Turtle is another widely used serialization format, which is designed specifically for RDF graphs (Figure 5). In contrary to OWL/XML, Turtle is more compact and easily readable for humans without requiring additional interpretation. Manchester syntax represents both human and machine readable serialization. While RDF/OWL represents named graphs, they can be easily expressed to humans in graphical representation (Figure 3).

```
:phenotype_diabetes_code rdf:type owl:Class ;
            owl:equivalentClass [ rdf:type owl:Class ;
                    owl:unionOf ( :diabdiag_gprd_code
                            :dm_gprd_code
                            :dm_hes_code
                            )
                    ] ;
            rdfs:subClassOf :code .
```

**Figure 5**. Example of Turtle serialization: part of the diabetes ontology and the definition of the diabetes code class used to denote terms from controlled clinical terminologies and their individual components. The class to hold diagnostic terms associated with diabetes diagnoses is a union between three classes of diagnostic terms specifically for primary and secondary care (e.g. diabdiag_gprd_code, dm_gprd_code, dm_hes_code).

*(ii) Implement set operations and relational algebra*

OWL natively supports set operations over defined classes and subjects. OWL allows us to define an operation of union, intersection and complement and to specify equivalent or disjoint classes and properties. SPARQL, a query language designed to query RDF graphs which grammar is similar to SQL supports relational algebra[11]. Figure 5 presents an example of OWL set operations of how the diabetes phenotype diagnostic terms are equivalent to a union of terms from three individual phenotype components.

*(iii)-(iv) Structured and temporal phenotype rule representation*

Nested logical structures can be created by merging individual structured RDF graphs. Comparative and aggregative operations are not implicitly supported by OWL itself, nevertheless SPARQL does support both as well as Boolean logic and negation and as a result, OWL constraints coupled with SPARQL and reasoners meet this criterion. As an example, the SPARQL query returning a set of patients with a type 1 diabetes diagnosis is illustrated in Figure 6. OWL/RDF does not directly support temporal relations, but since it natively provides a data-time datatype, the model for temporal relations could be designed and validity of the relation could be checked within individual SPARQL queries.

```
PREFIX rdf: <http://www.w3.org/1999/02/22-rdf-syntax-ns#>
PREFIX owl: <http://www.w3.org/2002/07/owl#>
PREFIX rdfs: <http://www.w3.org/2000/01/rdf-schema#>
PREFIX xsd: <http://www.w3.org/2001/XMLSchema#>
PREFIX clb: <https://www.caliberresearch.org/PhenotypeOntology#>
SELECT ?subject
        WHERE { ?subject rdf:type clb:subject_with_diabdiag_gprd_3_code .
        FILTER NOT EXISTS {?subject rdf:type clb:subject_with_type_unknown_diabetes .}
}
```

**Figure 6**. Example SPARQL query returning patients with a diabetes type 1 diagnosis.

*(v) Utilize standardized controlled clinical terminologies and facilitate reuse of value sets*

One of the main driving principles of ontologies is reusability. The majority of controlled clinical terminologies enable the unambiguous reference of particular terms by a unique identifier and RDF graphs are designed to reference these resource elements by an IRI. Additionally, many controlled clinical terminologies (e.g. Read, ICD-10, ICD-9) which do not natively support IRI can be obtained in a form of ontology on e.g. Bioportal (https://bioportal.bioontology.org/). Presence of such ontologies allows us not only to use a standardized controlled clinical vocabulary, but use them in the context of RDF natively. In our use-case, the diabetes phenotype algorithm uses Read terms for primary healthcare diagnosis and ICD-10 terms for hospital diagnosis. Currently, terms are added to the ontology as the literals however they can be also added as IRIs to their external resources. For example, the ICD-10 term *Insulin-dependent diabetes mellitus* (E10) has an IRI in the Bioportal ICD-10 terminology resource http://purl.bioontology.org/ontology/ICD10/E10.

*(vi) External software interfacing*

Our solution does not natively support an interface for external software algorithms or data components as it only contains an explicit specification without the executive components. However commonly used APIs like OWL API provide powerful tools to build such interfaces. Moreover, any external software algorithms implemented in an API proposed for OWL/RDF could be applied on the phenotype in OWL/RDF. NLP is not natively supported by OWL/RDF however, NLP rules and lexicons can be described by and extended by the phenotype ontology graph[12].

*(vii) Maintain backwards compatibility*

Backward compatibility on the level of versions is assured and OWL 2 is backward compatible to OWL 1. Additionally, as OWL/RDF are serialized into XML-based files, version control systems like Git (https://git-scm.com/) can potentially provide an effective manner to capture and control changes over time.

**Conclusion**

In this paper, we evaluated the use of SWT components such as OWL and RDF for storing computable representation of EHR-derived phenotyping algorithms and a proof of concept application using the OWL API and the Hermit reasoner was created. In our case study, we applied OWL in DL dialect on a deterministic diabetes EHR phenotype algorithm from the CALIBER resource. We propose a semi-automatic approach to constructing an ontology which serializes a phenotype description into a machine-readable XML-based file.

OWL/RDF has the potential to be a sufficient resource for storing phenotyping algorithms as it is versatile enough to meet most of the desiderata. The main significant limitation occurs in the case of supporting NLP methods for extracting clinical markers from unstructured text, which is not natively supported. On the other hand, NLP rules and lexicons for NLP engines could be specified in OWL and thus extend current phenotype ontology. Additionally, after ontology extension by structured EHRs in RDF, common reasoners like Hermit can automatically infer consequences and classify patients according to their diagnosis.

*Future work*

Future work includes evaluating the framework in additional phenotyping algorithms including complex phenotypes that make use of combinations of common EHR data such as diagnostic codes, lab measurements, prescription information or that rely on external data sources and integrate multimodal sources of clinical information (e.g. raw text found in patient charts or notes or medical images). Formal evaluation in terms of the algorithm implementation and it's accuracy in identifying patient cohorts should be undertaken and a comparison between other knowledge representation specifications and standards (e.g. the Guideline Interchange Format, GLIF[13]) would be very useful. As the automatic inferencing strongly depends on an algorithm logic specification which has to be added manually, a human-friendly interface should be implemented and formally evaluated to enable researchers to specify algorithmic logic easily and through a graphical user interface. Researchers should be allowed to specify algorithm logic (i.e., to create rules between classes) and ideally queries using SPARQL even without intimate knowledge of a programming language. Development of the user-friendly interface addresses one of many shortcomings of the semantic web[14] and thus detailed investigation of these shortcomings (e.g. existence of competitive ontologies, performance loss for large datasets) and their influence on this work is highly beneficial for the future progress.

**References**

1. S. Denaxas and K. Morley, "Big biomedical data and cardiovascular disease research: opportunities and challenges". European Heart Journal: Quality of Care and Clinical Outcomes, vol. 1, 2015, pp. 9-16, doi: 10.1093/ehjqcco/qcv005.
2. W. Wei, J. C. Denny, "Extracting research-quality phenotypes from EHR to support precision medicine". Genome Medicine, 2015, doi: 10.1186/s13073-015-0166-y.
3. K. Morley, J. Wallace, S. Denaxas, et al., "Defining Disease Phenotypes Using National Linked Electronic Health Records: A Case Study of Atrial Fibrillation". PLOS ONE, 2014, doi: 10.1371/journal.pone.0110900.
4. C. G. Chute, J. Pathak, G. K. Savova, et al., "The SHARPn Project on Secondary Use of Electronic Medical Record Data: Progress, Plans, and Possibilities". AMIA Annual Symposium, 2011, pp. 248-256.
5. C. McCarty, R. Chiholm, C. Chute, et al., "The eMERGE Network: A consortium of biorepositories linked to electronic medical records for conducting genomic studies". BMC Medical Genomics, vol. 4, 2011, doi: 10.1186/1755-8794-4-13.
6. G. Hripcsak, D. Albers, "Next Generation Phenotyping of EHR". JAMIA, 2013, vol. 20, pp. 117-121, doi: 10.1136/amiajnl-2012-001145
7. J. Pathak, R. Kiefer, C. Chute, "Using Semantic Web Technologies for Cohort Identification from Electronic Health Records for Clinical Research". AMIA Jt Summits Transl Sci Proc, 2012, pp. 10-19.
8. S. Denaxas, J. George, E. Herrett, et al., "Data resource profile: cardiovascular disease research using linked bespoke studies and electronic health records (CALIBER)". Int J Epidemiol, vol. 41, 2012, pp. 1625-38, doi: 10.1093/ije/dys188

9. S. Shah, C. Langenberg, E. Rapsomaniki, S. Denaxas, M. Pujades-Rodriguez, C. Gale, J, Deanfield, L. Smeeth, A. Timmis, H. Hemingway, "Type 2 diabetes and the incidence of cardiovascular diseases: a cohost study of 1.9m people". Lancet Diabetes Endocrinol, vol. 3, 2015, pp. 105-13, doi: 10.1016/S2213-8587(14)70219-0
10. H. Mo, W. Thompson, L. Rasmussen, et al., "Desiderata for computable representations of EHR-driven phenotype algorithms". JAMIA, vol. 22, 2015, pp. 1220-30, doi: 10.1093/jamia/ocv112
11. R. Cyganiak, "A relational algebra for SPARQL". Digital Media Systems Laboratory, 2005.
12. A. Toral, M. Monachini, "SimpleOWL: a generative lexicon ontology for NLP and the semantic web". Workshop on Cooperative Construction of Linguistic Knowledge Bases, 2007.
13. A. A. Boxwala, M. Peleg, S. Tu, et al., "GLIF3: a representation format for sharable computer-interpretable clinical practice guidelines". Journal of Biomedical Informatics, vol. 37, 2004, pp. 147-161, doi: 10.1016/j.jbi.2004.04.002
14. X. Zenuni, B. Raufi, F. Ismaili, J. Ajdari, "State of the Art of Semantic Web for Healthcare". Procedia-Social and Behavioral Science, vol. 195, 2015, pp. 1990-1998, doi: 10.1016/j.sbspro.2015.06.213